\documentclass[runningheads]{llncs}
\usepackage{eccv}
\usepackage{graphicx}

\usepackage{eccvabbrv}

\usepackage{graphicx}
\usepackage{booktabs}
\usepackage{xcolor}
\usepackage{multirow}
\usepackage{amsmath}
\usepackage{graphicx}
\usepackage{subcaption}
\usepackage{hyperref}

\usepackage{orcidlink}
\usepackage{booktabs}
\usepackage{diagbox}
\usepackage[linesnumbered,ruled,vlined]{algorithm2e}
\usepackage{amssymb}
\usepackage{amssymb}
\usepackage{pifont}
\usepackage{hyperref}

\usepackage[accsupp]{axessibility}  

\begin{document}
%
\title{Structured Analysis and Comparison of Alphabets in Historical Handwritten Ciphers}

\titlerunning{Structured Analysis of
Alphabets in Historical Handwritten Ciphers}
%


\author{Martín Méndez\orcidID{0009-0009-6722-4942} \and
Pau Torras \orcidID{0000-0003-0327-9046} \and
Adrià Molina \orcidID{0000-0003-0167-8756}  \and Jialuo Chen\orcidID{0000-0002-7808-6567} \and Oriol Ramos-Terrades \orcidID{0000-0002-3333-8812} \and Alicia Fornés \orcidID{0000-0002-9692-5336}}%
\authorrunning{M. Méndez \textit{et al.}}
%
\institute{Computer Vision Center,  
Computer Science Department \\
Universitat Autònoma de Barcelona, Catalunya \\
\email{martin.mendez01@estudiant.upf.edu,\\ \{ptorras, amolina, jchen, oriolrt, afornes\}@cvc.uab.cat} }
\maketitle              
\begin{abstract}

Historical ciphered manuscripts are documents that were typically used in sensitive communications within military and diplomatic contexts or among members of secret societies. These secret messages were concealed by inventing a method of writing employing symbols from diverse sources such as digits, alchemy signs and Latin or Greek characters. When studying a new, unseen cipher, the automatic search and grouping of ciphers with a similar alphabet can aid the scholar in its transcription and cryptanalysis because it indicates a probability that the underlying cipher is similar. In this study, we address this need by proposing the CSI metric, a novel way of comparing pairs of ciphered documents. We assess their effectiveness in an unsupervised clustering scenario utilising visual features, including SIFT, pre-trained learnt embeddings, and OCR descriptors.
\end{abstract}
\section{Introduction}
\label{sec:intro}

A ciphered manuscript is a unique type of historical document containing secret messages, commonly used in military or diplomatic correspondence, by secret societies, or in diaries and private letters. Historians estimate that about 1\% of archival material consists of encrypted documents, which means that a very significant -- and oftentimes, critical -- part of our common historical past is inaccessible \cite{megyesi2020decryption}. 

These historical ciphers are usually designed around the same fundamental principles. The sender and the receiver agree on a secret encoding method to conceal a message from undesired readers (in contrast to modern cryptography techniques where the method is usually publicly known and only the keys are secret). The resulting ciphers are constructed by assembling a vocabulary of symbols, either invented or borrowed from a wide range of alphabets and graphical sources, with complex correspondences to those in the plain text. More specifically, it is common to find ciphers with digits, alphabetical characters (e.g., Latin or Greek letters), punctuation marks, diacritics and other graphical signs such as Zodiac or alchemy symbols. Additionally, transposition and substitution of characters are applied to further strengthen the cipher. Unsurprisingly, this aspect makes their transcription and cryptanalysis extremely difficult, particularly when the documents in question comprise a small corpus of text.

Hence, when studying a newly encountered cipher, the first step is to identify its alphabet to see whether similar ones have been studied before. If a cipher employs a similar set of symbols, it is likely that it also employs a similar key or that it uses the same plain text language. This identification task becomes challenging when the alphabet contains invented symbols or when the symbols are arranged in a complex or overlapping layout. Despite advancements in document analysis systems for classifying scripts without supervision\cite{nicolau2016visual,christlein2017unsupervised,keserwani2019zero,shi2016script},there is still a scarcity of literature on such methods that are applicable to ciphers \cite{chen2021unsupervised}. Additionally, since ciphered documents are rarely transcribed, any image processing methods should avoid the need of annotated datasets, which limits the applicability of supervised Deep Learning methodologies.

For these reasons, in this work we propose a novel unsupervised technique for the automatic identification and comparison of cipher alphabets without requiring annotated data. Concretely, we propose the Cipher Similarity Index (CSI) metric, a new way of quantifying the similarity of pairs of ciphers that improves upon previous literature \cite{chen2021unsupervised} by generating metric values less reliant on either the source of extracted features. In addition, we perform a thorough comparison between unstructured and structured methods for alphabet analysis, discussing their performance on various types of cipher manuscripts.

\section{Related Work}
\label{sec:sota}

The recognition of cipher manuscripts is challenging, not only because of the variability in the handwriting styles, but also due to the often unknown cipher alphabet. Nevertheless, there have been several works that have focused on the recognition of cipher documents, either using clustering techniques \cite{yin2019decipherment}, recurrent neural networks \cite{souibgui2023evaluation} and few-shot learning \cite{souibgui2022few}. However, most of these methods require the intervention of the scholar to identify the set of unique symbols of the cipher or to provide some transcriptions for training. 

In the domain of script analysis, there's a noticeable shift towards unsupervised or zero-shot inference techniques. Initial findings reported in \cite{chen2021unsupervised} indicated that clustering methods like k-means could yield significant performance gains without the need for supervision. This approach contrasts with prevailing literature \cite{leierzopf2021massive,leierzopf2021detection}, which heavily relies on supervision and text features.

The reliance on text features can be problematic in scenarios where annotations are scarce. In such cases, existing research provides compelling evidence of the feasibility of operating at the pixel level, whether through hand-crafted features \cite{nicolau2016visual} or learned descriptors, such as those obtained from feature learning with preliminary pseudo-labels \cite{christlein2017unsupervised}, zero-shot \cite{keserwani2019zero}, and discriminative \cite{shi2016script} approaches. 

Despite recent advancements in zero-shot and representation learning for script identification, there remains a notable interest in the potential of structured methods for outlier detection, particularly in classifying out-of-distribution data. For instance, Ozaki \etal \cite{ozaki2011using} utilized a mutual-knn graph over textual data to detect anomalies, achieving significant performance gains. This finding has been further corroborated in the context of text processing \cite{liu2012noisy} and, more recently, in structured topological data \cite{terrades2022flexible}. In both cases, authors demonstrate the advantages of mutual-knn approaches in localising out-of-domain and outlying data.

In summary, it is evident from the literature in cipher and script classification that there is a growing need to leverage pixel-based methods to reduce the reliance on annotation resources. While preliminary findings by Chen \etal \cite{chen2021unsupervised} touch upon this aspect, there remains a necessity to comprehensively compare the effectiveness of hand-crafted features \cite{nicolau2016visual} with previously learned descriptors (zero-shot, \cite{chen2021unsupervised,keserwani2019zero,shi2016script}) in a structured manner \cite{liu2012noisy,ozaki2011using}.

\section{Methodology} \label{sec:methodology}

In this section we introduce our proposed method to analyse the similarity between ciphered documents using unsupervised techniques. The main intuition behind it is that images of different versions of the same symbol should be close from one another in the feature space. Therefore, ciphers containing a similar subset of symbols should be difficult to separate completely because their samples should lie close in the feature space. This method assumes that a symbol-level segmentation is possible and that features are extracted from each symbol.

\subsection{Baseline}
\label{sec:arch2}

As baseline, the unsupervised method described in Chen \etal \cite{chen2021unsupervised} is used, which makes the same assumptions. In this method, symbols are segmented from the document and a feature vector is computed for each. After this, the $k$-means clustering algorithm is applied in order to group visually similar symbols. To ensure a fair comparison between different cipher alphabets, it is important to have balanced data. For this reason, an equal number of symbols from each manuscript is taken.

Once clusters are obtained for each pair of ciphers, the method analyses their similarity. This is done by examining the composition of each cluster and determining whether its elements belong exclusively to either or both. Clusters with predominantly single-cipher symbols indicate different alphabets, while clusters with a mixture of symbols from both ciphers suggest higher similarity between the two ciphers. This similarity is quantified numerically by computing the ratio of mixed clusters against the total number of clusters.

\subsection{Cipher Similarity Index (CSI)}
\label{sec:arch}

\begin{figure}
    \centering
    \includegraphics[width=0.9\textwidth]{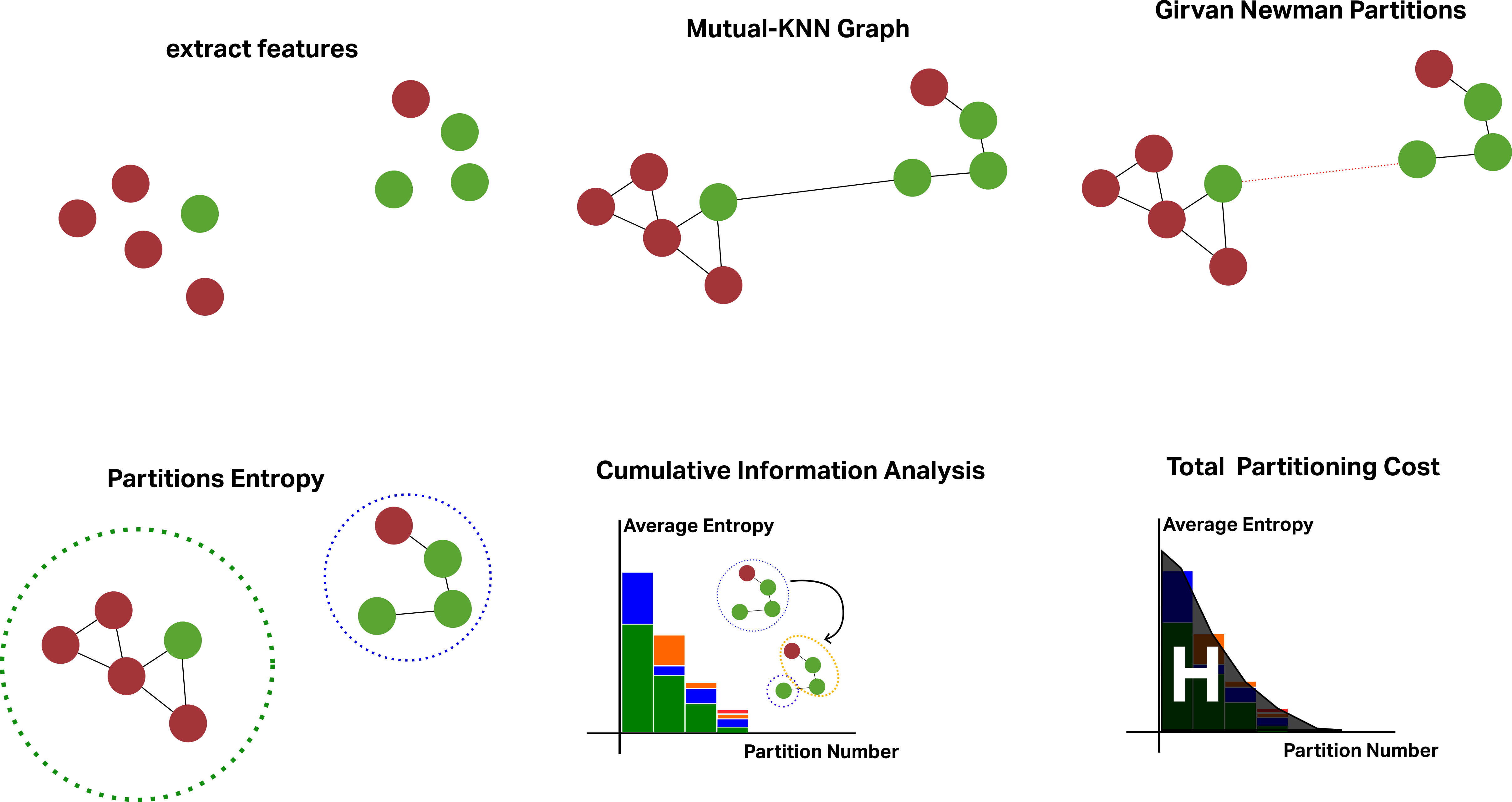}
    \caption{Summary of the graph-based method for clustering of cyphers.}
    \label{fig:method_entropy}
\end{figure}

Building upon previous literature \cite{liu2012noisy,ozaki2011using}, we aim to integrate a structured approach, summarised in Figure \ref{fig:method_entropy}, on the baseline introduced at \ref{sec:arch2}. In essence, what is expected from the CSI metric is to palliate the effect of the choice of clusters during \textit{k}-means and the source of features. The idea is to avoid considering loosely coupled homogeneous groups of samples as large mixed clusters.

The method starts with the same pre-computed features as the baseline method, from which a Mutual-KNN graph is generated. In contrast to KNN graphs, which can generate links between samples regardless of their distance, mutuality has proven to reduce noise in outlier classification regimes \cite{liu2012noisy,brito1997connectivity}. After constructing the graph, the Girvan-Newman algorithm \cite{girvan2002community} is applied to it, which iteratively removes its longest edges. Each iteration produces one additional cluster in the graph that originates from splitting 
the most loosely coupled community at the previous time-step.

Subsequently, $M$ partitioning steps are executed to progressively refine the communities, resulting in a vector ranging from $N$ to $N+M$ clusters. At every iteration, the Shannon Entropy (Eq. \ref{eq:entropy}) is computed on each cluster of the graph in order to quantify the heterogeneity of the labels within it (i.e. whether that cluster contains samples originating from one or multiple source documents). The Shannon Entropy $H_c(X)$ of cluster $c$ on a set of distinct documents $X$ is computed from the probability $p_c(x)$ of having a sample of class $x$ in cluster $c$ and defined as
\begin{equation}
    H_c(X) = - \sum_{x \in X} p_c(x) \log p_c(x).
    \label{eq:entropy}
\end{equation}
The maximal value is achieved when all classes are equally probable. In other words, the entropy is maximal when clusters contain heterogeneous sets of labels. Analogously, the entropy is zero when all labels belong to the same class, indicating homogeneity. This is extended for the whole graph at a given iteration by computing the weighted average of all entropy values considering the number of samples present in each cluster.

\begin{figure}[htb]
    \centering
    \includegraphics[width=0.85\textwidth]{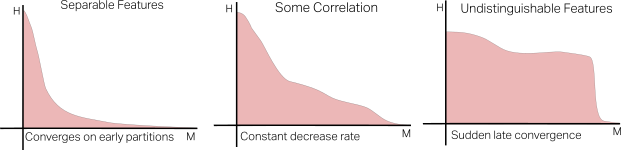}
    \caption{Dissimilar datasets often exhibit delayed convergence in the entropy partitioning scheme. Hence, the area under the curve serves as a metric for assessing the separability of features.}
    \label{fig:conv}
\end{figure}

Performing this calculation on every iteration of the Girvan-Newman algorithm provides a measurement on the degree of separation of the classes of the data. If the overall entropy decreases sharply at the beginning of the iterative process it means that the first few edges are the ones that connect communities belonging to different documents (see Fig. \ref{fig:conv}). Therefore, the data is easily separable and the documents are dissimilar. Otherwise, if the clusters are not homogeneous until the very end of the iterative process it means that the documents being analysed are tightly coupled and therefore very similar. This idea can be seen in Figure \ref{fig:conv} and can be summarised into a global similarity metric by computing the area under the curve of the global entropy per iteration. We have named this metric \textbf{Cipher Similarity Index (CSI).}

An important aspect to consider when computing the CSI is that the maximal entropy depends on the prior distribution of the document samples. If the classes are not equiprobable \textit{a priori} then the metric is not bounded to the range of 0 to 1, but rather the range defined by the entropy of the prior distribution. This can be sidestepped by normalising with respect to the maximum value of the entropy for the measured set of samples or by sampling a fixed number of elements from each document, which can be exploited to reduce the computational complexity of the algorithm.

\section{Experimental Setup}
In the experiments, we hypothesise that labelled data is not available. So, we separate samples from different source documents assuming that each document contains a unique alphabet. If two documents' scripts share a significant amount of symbols, they should be considered similar using our metric. To assess this hypothesis, we have compiled a collection of manuscripts, presented in Section \ref{subsec:data}. These contain both unique and shared alphabets with differing properties to provide some insight on the feasibility of our proposed methodology.

As mentioned in Section \ref{sec:methodology}, the baseline and our proposed method require the segmentation of symbols for each ciphered document. The pre-processing is described in \ref{subsec:preprocess}. Moreover, they both require extraction of features on the symbol-level images, which are described in \ref{subsec:extractors}.

\subsection{Datasets} \label{subsec:data}

\begin{figure}[ht]
    \begin{center}
    \begin{tabular}{c}

    \hline
    \multicolumn{1}{|l|}{\includegraphics[width=0.8\columnwidth, height=0.5cm]{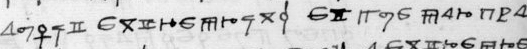}}\\
    \hline
    \small Bnf Cipher\\
    
    \hline
    \multicolumn{1}{|l|}{\includegraphics[width=0.8\columnwidth, height=0.5cm]{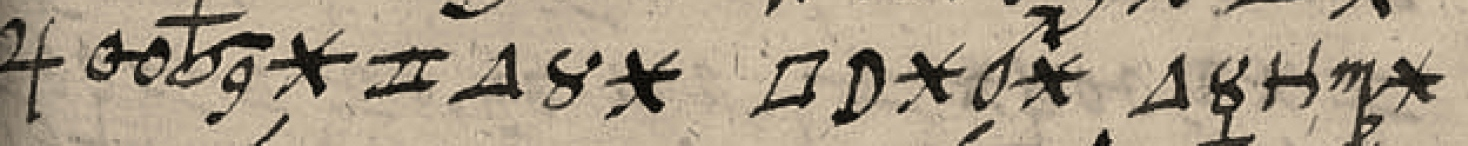}}\\
    \hline
    \small Borg Cipher\\
    
    \hline  
    \multicolumn{1}{|l|}{\includegraphics[width=0.8\columnwidth, height=0.5cm]{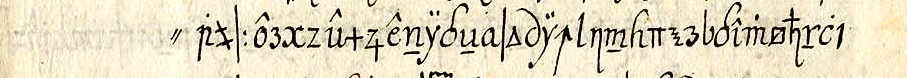}}\\
    \hline
    \small Copiale Cipher\\
    
    \hline  
    \multicolumn{1}{|l|}{\includegraphics[width=0.8\columnwidth, height=0.5cm]{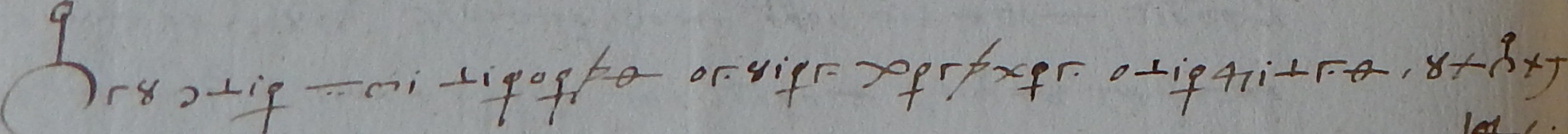}}\\
    \hline
    \small MaximilianII Cipher\\
    
    \hline  
    \multicolumn{1}{|l|}{\includegraphics[width=0.8\columnwidth, height=0.5cm]{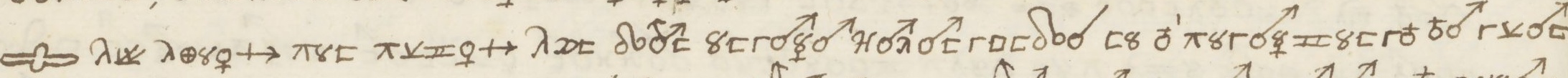}}\\
    \hline
    \small Ramanacoil Cipher\\

    \hline  
    \multicolumn{1}{|l|}{\includegraphics[width=0.8\columnwidth, height=0.5cm]{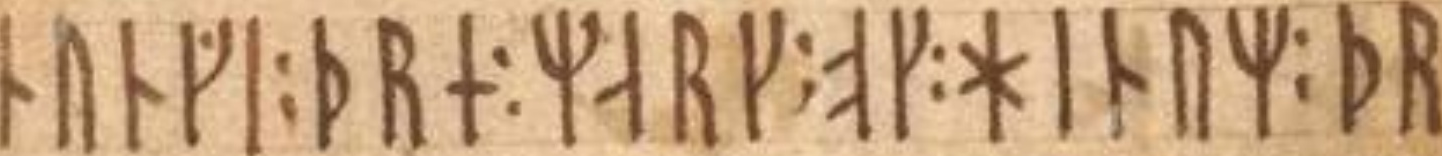}}\\
    \hline
    \small Codex Runicus\\
    
    \hline  
    \multicolumn{1}{|l|}{\includegraphics[width=0.8\columnwidth, height=0.5cm]{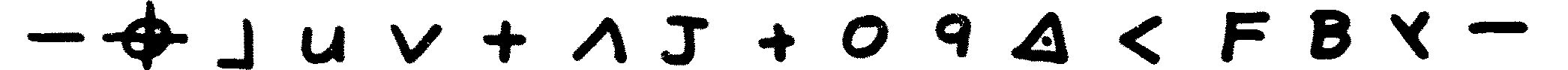}}\\
    \hline
    \small Zodiac Cipher\\

    \hline  
    \multicolumn{1}{|l|}{\includegraphics[width=0.8\columnwidth, height=0.5cm]{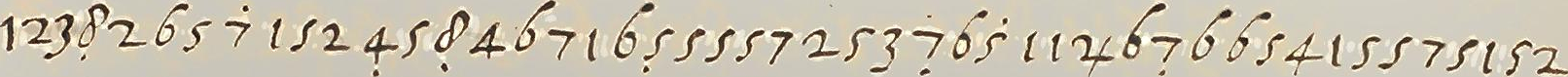}}\\
    \hline
    \small Vatican 1 Cipher\\
    
    \hline  
    \multicolumn{1}{|l|}{\includegraphics[width=0.8\columnwidth, height=0.5cm]{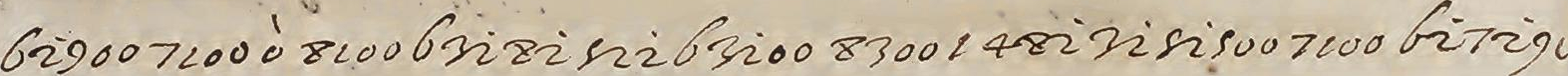}}\\
    \hline
    \small Vatican 2 Cipher\\
    
    \hline  
    \multicolumn{1}{|l|}{\includegraphics[width=0.8\columnwidth, height=0.5cm]{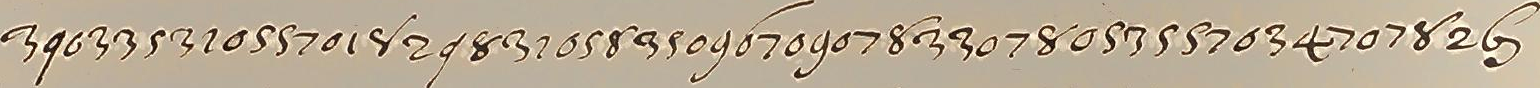}}\\
    \hline
    \small Vatican 3 Cipher\\

    \hline  
    \multicolumn{1}{|l|}{\includegraphics[width=0.8\columnwidth, height=0.5cm]{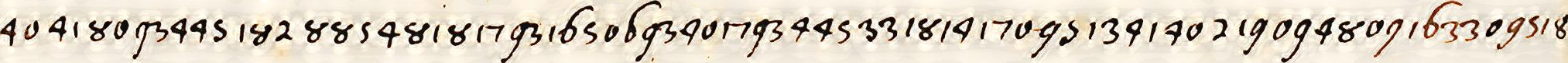}}\\
    \hline
    \small Vatican 6 Cipher\\

    \hline  
    \multicolumn{1}{|l|}{\includegraphics[width=0.8\columnwidth, height=0.5cm]{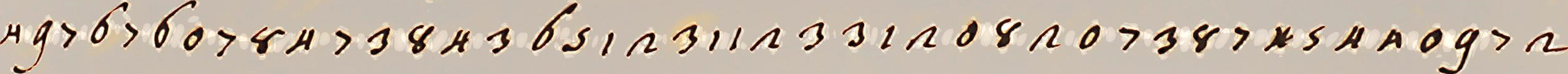}}\\
    \hline
    \small Vatican 7 Cipher\\
    
    \end{tabular}
    \end{center}
    \caption{Samples from the manuscripts employed throughout this study.}
    \label{fig:datalines}
\end{figure}


Different manuscripts have been used in the experiments. Some of them contain a unique alphabet, whereas others share the same alphabet but with different handwriting styles (e.g. digits in Vatican ciphers). Examples of each manuscript are shown in Figure \ref{fig:datalines}. The documents are described next:

\textbf{The BNF cipher} is a corpus of ciphered letters written by various French nobles around the 16th century, kepth in the Bibliotheque Nationale de France. It contains about 37 distinct types of graphical symbols, generally non-touching. The plaintext - before encryption - is French.

\textbf{The Borg cipher}\footnote{https://www.su.se/english/research/research-projects/decipherment-of-historical-manuscripts/the-borg-cipher-1.688283} is a long encrypted manuscript from the 17th century. All the manuscript is encoded, except the first and last two pages. The cipher consists of 34 different symbols, generally touching, and combining graphic signs, Latin letters and diacritics.  The plaintext language - before encryption - is Latin.

\textbf{The Copiale cipher} \footnote{https://www.su.se/english/research/research-projects/decipherment-of-historical-manuscripts/the-copiale-cipher-1.688288} is a long encrypted manuscript from the mid 18th century. The cipher employs 100 different symbols, using characters from the Latin and Greek alphabets and ideograms that represent special entities. The plaintext language - before encryption - is German.

\textbf{The Maximilian II cipher} \cite{nils2022deciphering} consists in encrypted diplomatic letters sent by Maximilian II in 1575. It uses different graphic signs and can be described as a mixture of astrological signs, Greek letters, and esoteric symbols. 

\textbf{The Ramanacoil cipher} \cite{dinnissen2021island} is a manuscript from 1674, kept in the National Archives of the Netherlands. It employs 24 unique (non-touching) symbols for the Latin alphabet (but without V and J), additional special symbols for double letters (EE, FF, LL, OO, and PP), and special symbols for seven important words (e.g. “Ramanacoil”). The plaintext language - before encryption - is Dutch.

\textbf{The Codex Runicus} \cite{peratello2020codex} is a medieval manuscript from the 14th century containing the oldest preserved Nordic provincial law, written using the runic alphabet. It was written on 100 parchment folios, totaling 202 pages. This manuscript is not a cipher, but it has been included to show that our methodology can be extended to any script system, not only cipher alphabets.

\textbf{The Vatican ciphers} are a collection of documents found at the Secret Archive of the Vatican. The selection presented here (identified with the numbers 1, 2, 3, 6, 7) span various centuries and handwriting styles. These Vatican ciphers are generally written using 76 different symbols, based mostly on digits with multiple diacritics.

\textbf{The Zodiac cipher}\footnote{https://www.dcode.fr/zodiac-killer-cipher} is the generic name given to a series of modern ciphers written by a serial killer based in the San Francisco Bay, United States between the 1960s and 1970s. The plaintext - before encryption - is English.

\subsection{Preprocessing}  \label{subsec:preprocess}


Segmented symbols are the input to all models. We have used the same unsupervised symbol segmentation algorithm as in \cite{chen2021unsupervised}. First, the document is binarised using Sauvola thresholding \cite{opencv_thresholding_tutorial}. Text lines are then segmented by summing all columns of the image matrix and detecting peaks with the peakutils library\cite{lucas_hermann_negri_2017_887917}. Once every text line is isolated, symbols are segmented by finding connected components. The final symbol reconstruction requires a final step of aggregation of close objects, as there are symbols formed by the composition of smaller semantic units (e.g. dotted symbols). In Table \ref{tab:dataset_samples}, the final number of segmented symbols per document is shown.

\begin{table}[htb]
\centering
\caption{Number of segmented symbols for each dataset.}
\label{tab:dataset_samples}
\resizebox{\columnwidth}{!}{%
\begin{tabular}{@{}cccccccccccc@{}}
\toprule
BNF  & Borg & Copiale & Maxim. & Raman. & Runicus & Vat. 1 & Vat. 2 & Vat. 3 & Vat. 6 & Vat. 7 & Zodiac \\ \midrule
2988 & 1140 & 3472    & 3939          & 12226      & 3206    & 1831     & 6111     & 2391     & 4407     & 4522     & 593    \\ \bottomrule
\end{tabular}%
}
\end{table}

\begin{figure}
    \centering
    \begin{subfigure}{.5\textwidth}
        \centering
        \includegraphics[width=0.9\linewidth]{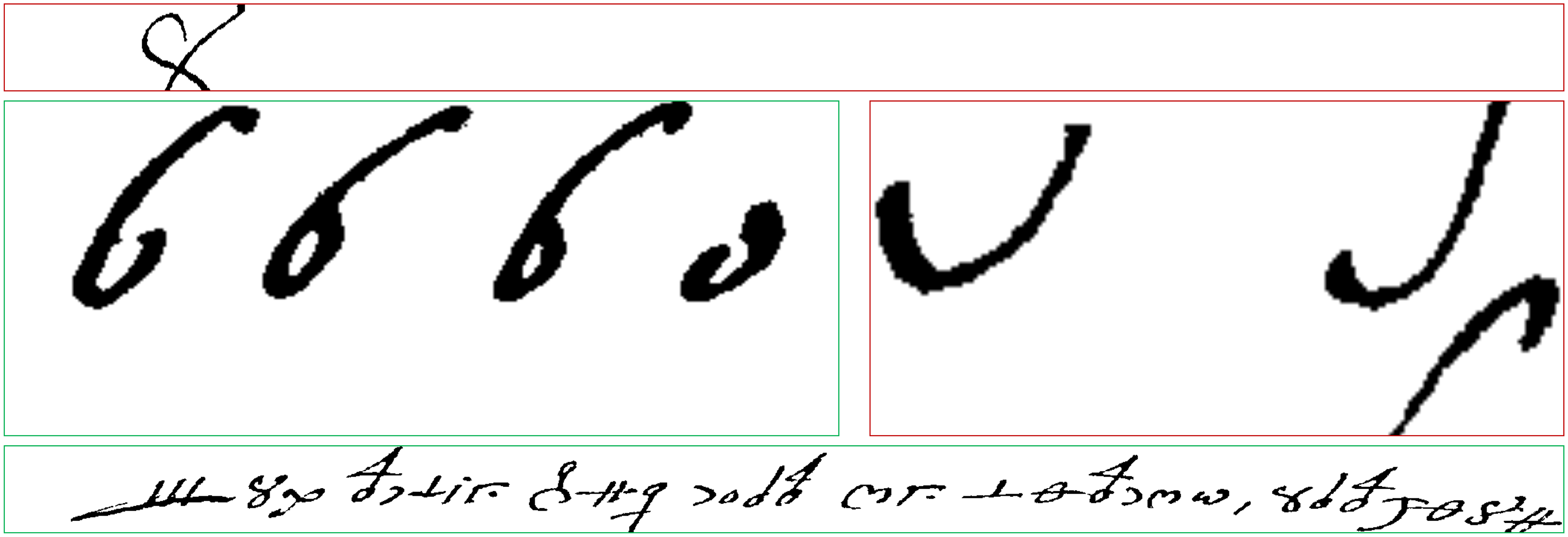}
        \caption{}
        \label{fig:sub1}
    \end{subfigure}%
    \begin{subfigure}{.4\textwidth}
        \centering
        \includegraphics[width=.4\linewidth]{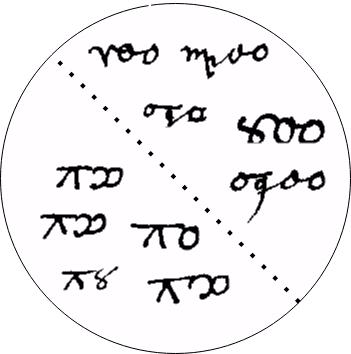}
        \caption{}
        \label{fig:sub2}
    \end{subfigure}%
    \caption{Examples of incorrectly segmented symbols. a) Segmentation errors due to line detection failure (red) and correct segmentation (green). b) Cluster with wrongly segmented symbols.}
    \label{fig:qual_segm}
\end{figure}

It should be noted that, since this method is completely unsupervised, there can be  segmentation mistakes. Figure \ref{fig:qual_segm} shows some prototypical examples of errors during this process.

\subsection{Feature Extractors} \label{subsec:extractors}

To discuss the usability of learnt vs handcrafted descriptors, we run experiments using different features extracted from the segmented symbols, concretely:

\textbf{SIFT.}
First, we utilise the SIFT\cite{lowe2004distinctive} (Scale-Invariant Feature Transform) descriptor to extract distinctive local features from images. This method involves detecting key points in the image that are invariant to scale and rotation, followed by computing descriptors for these points based on gradients in the surrounding area. The way we have extracted the features is by resizing the image to be large enough to extract features from it (in our case, to 64x64), and applying a 4x4 grid of keypoints based on the width and height of the image from which features are to be extracted. We used the SIFT implementation from \texttt{OpenCV}\footnote{https://docs.opencv.org/4.5.5/da/df5/tutorial\_py\_sift\_intro.html} to generate these descriptors.

\textbf{VGG16.}
Secondly, we extract features using a 16-layers deep convolutional neural network, namely VGG16\cite{simonyan2015deep}. This architecture is a usual way-to-go for practitioners, as the learnt features in narrow layers tend to be generic enough to be useful in describing out-of-domain images \cite{zheng2017sift}. More precisely, we are using a \texttt{torchvision}\footnote{github.com/pytorch/vision} VGG-16bn implementation pretrained on imagenet\cite{deng2009imagenet}.

\textbf{CLIP.}
Alternatively, we take advantage of CLIP\cite{radford2021learning} trained on Laion-2b\cite{schuhmann2022laion5b} as a state-of-the-art visual encoder. We use the \texttt{open-clip}\cite{openclip_github} implementation of a ViT-B/32\cite{dosovitskiy2021image}.

\textbf{OCR.}
Lastly, it is important to take into consideration the textual nature of the data at hand. Therefore, we train a CTC\cite{graves2006connectionist} ViT\cite{dosovitskiy2021image} with a transformer decoder \cite{vaswani2017attention} to obtain transcriptions of scene and document images (\textbf{OCR-Generic}) by training on the HierText\cite{long2022towards} dataset. We also provide results with the same architecture, but specialised on handwritten datasets (\textbf{OCR-Handwritten}) such as \cite{fischer2011transcription}, \cite{fischer2012lexicon} and \cite{romero2013esposalles}. In both cases, the descriptors are obtained by aggregation of the tokens produced by the encoder via max-pooling.

For computational efficiency, the computed features are passed through PCA to reduce their dimensionality to 50 before being fed to our methods.

\section{Experimental Results}

In this Section, we present the results obtained using the Baseline method and our proposed CSI method.

\subsection{Baseline Results}

\begin{figure}[ht]
    \centering
    \includegraphics[width=\textwidth]{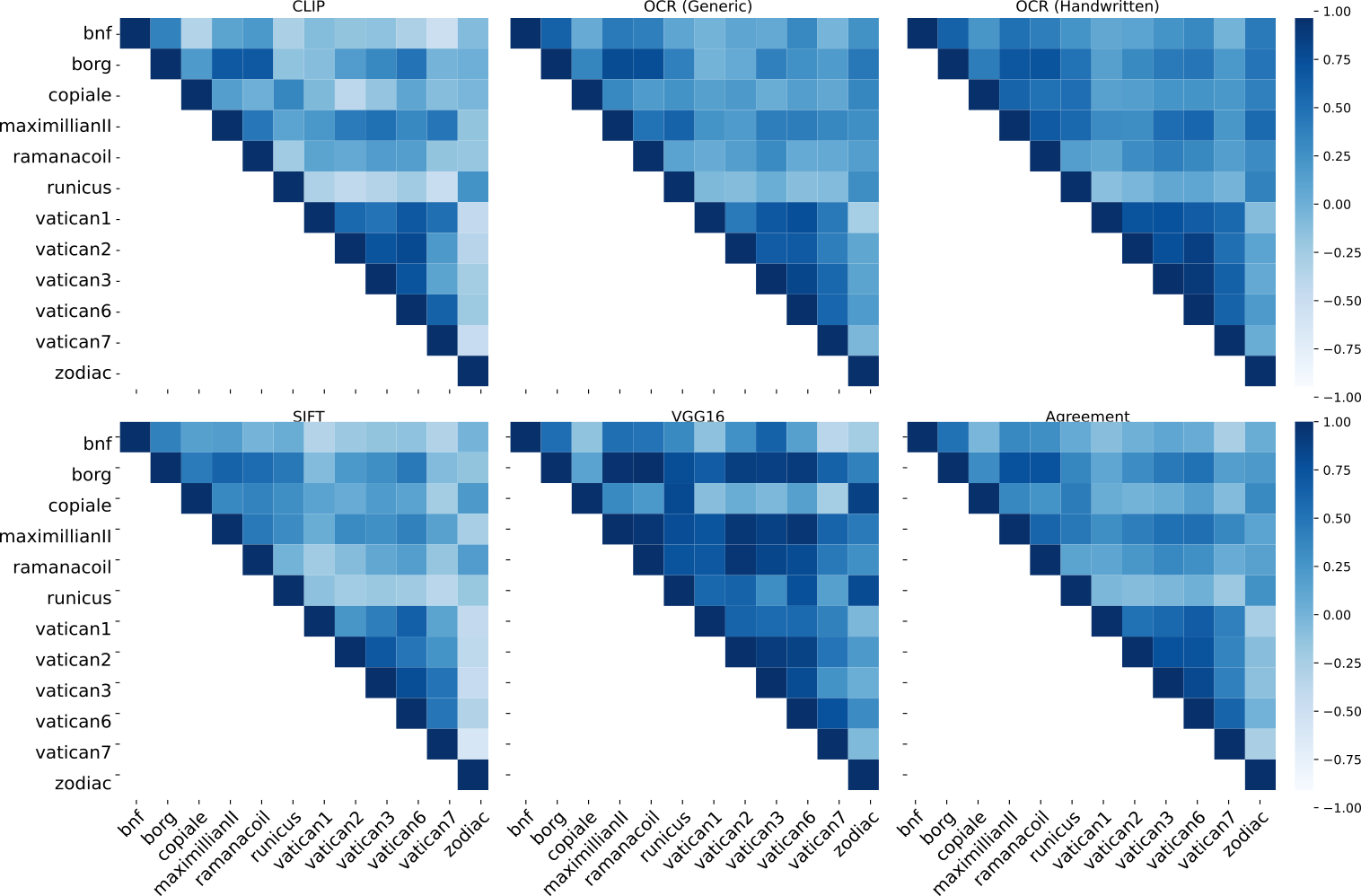}
    \caption{Baseline results between pairs of ciphered documents using features from a CLIP model, an OCR model trained on generic text, an OCR model trained on handwritten text, a VGG16 and SIFT. The last cell shows the degree of agreement between all feature sources.}
    \label{fig:baseline_results}
\end{figure}

Results on the Baseline method are shown in Figure \ref{fig:baseline_results} and are obtained by averaging the cluster similarity percentages obtained on 4 runs after sampling a random subset of 500 feature vectors from each cipher. This can be computed fairly quickly -- with each run taking around 20 to 50 seconds in a commercial CPU with our implementation -- unlike with the full set of samples, which for certain pairs of documents can take hours. The agreement score is defined as
\begin{equation}
    \text{AGR} = \left( \prod_{\forall q \in Q} \text{Baseline}(q)\right)^{1/|Q|}
\end{equation}
where $Q$ is the set of feature sources (VGG, SIFT, etc.). It should be interpreted as an average of values produced using CSI on all feature sources.

The main behavioural aspects to outline are the fact that most feature sources result in Vatican documents being assumed similar (which makes sense since Vatican documents mainly use digits), which confirms the intuitions and assumptions about the method. Nevertheless, some of its shortcomings are the general homogeneity of similarity values and the failure case using VGG16 features, where all documents are deemed similar. Our re-run experiments resonate with those of the original publication \cite{chen2021unsupervised}.

\subsection{Cipher Similarity Index}

Experiments on CSI are performed using the same process of averaging 4 runs using a random subset of 500 feature vectors from each cipher. Using CSI each run takes between 20 and 30 seconds, depending on the pair of ciphers being studied.

\begin{figure}[ht]
    \centering
    \includegraphics[width=\textwidth]{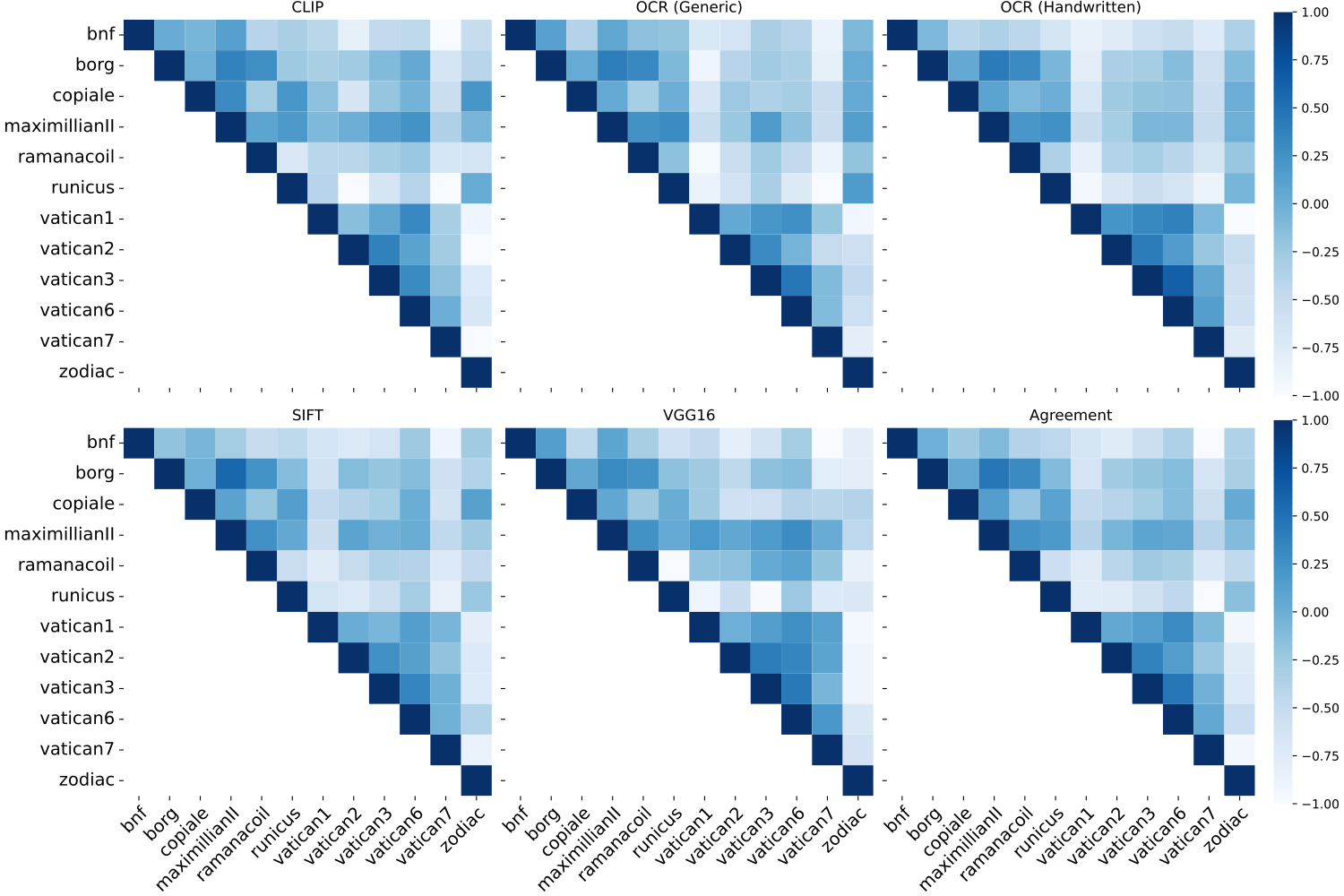}
    \caption{Normalised CSI values between pairs of ciphered documents using features from a CLIP model, an OCR model trained on generic text, an OCR model trained on handwritten text, a VGG16 and SIFT. The last cell shows the degree of agreement between all feature sources.}
    \label{fig:csi_results}
\end{figure}

In Figure \ref{fig:csi_results} the results for all feature extraction methods using the CSI metric are shown. The values are normalised for ease of comparison as $\text{CSI} - \mu / \sigma$ where $\mu$ and $\sigma$ are the mean and standard deviation of all values of CSI for a single feature. The same definition of agreement score is used.

The main conclusion that can be drawn from Figure \ref{fig:csi_results} is that the source of the features does not seem to have a strong effect on the final value of the metric. As it can be seen from the agreement score, the algorithm operates similarly on all feature spaces, which indicates that all feature extraction methods result in each document being placed within the same relative distances from each other. This is fairly indicative that features extracted from each segmented symbol are also representative of their origin, which aligns with our initial hypotheses.

\begin{figure}[ht]
    \centering
    \includegraphics[width=0.8\textwidth]{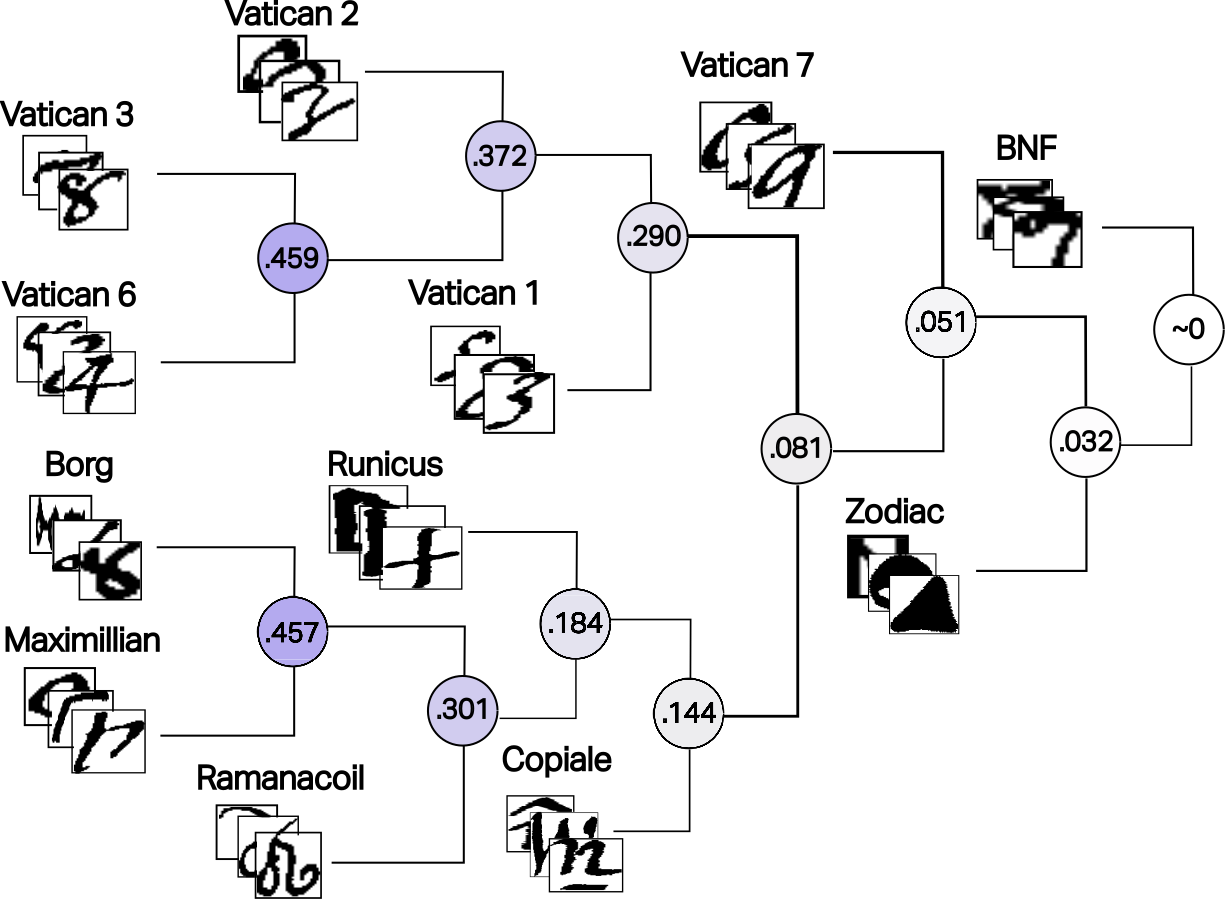}
    \caption{Qualitative results depicting the agglomerative pairing of the two most similar ciphers based on the normalised aggregate CSI score. It is noteworthy how digit-based ciphers (e.g., Vatican) are distinctly separated from symbol-based ones (e.g., Borg, Ramanacoil). Both modalities converge in later stages.}
    \label{fig:pairing_qualitative}
\end{figure}

A second conclusion is that CSI seems to behave as expected for those ciphers that are known to have shared or partly shared alphabets. In particular, all documents encoded using the Vatican cipher appear to be similar according to our metrics, which correlates with the fact that all these ciphers use digits as their base alphabet. An interesting point is that this effect seems to be more pronounced for OCR features, possibly because digits are in-domain for these methods whereas other symbols are not. For combinations of other pairs of similar ciphers a more definite answer is difficult without the criterion of an expert paleographer. In Figure \ref{fig:pairing_qualitative} some qualitative results are given for the highest similarity values for each cipher for illustration purposes. Using the results from Vatican it can also be ascertained that this system seems to be partially robust to variations in handwriting style because of the existence of significant differences in the source documents (character slanting, spacing, cleartext, etc.), but the extent of this is difficult to evaluate from the CSI metric alone.

When comparing with the results using the Baseline method, the most evident difference is the fact that CSI values have a broader relative range, implying the output values are more informative of similarity. Moreover, there is less variance among the source of features. The same overall patterns in similarity are shared between both methods, indicating that they are both drawing the same conclusions about the relationships between pairs of documents.

\subsection{A different approach on out-of-domain identification}

In this Subsection we propose another way of computing cipher similarities using a classification model trained on a specific subset of documents. The purpose is to check whether similar results could be obtained through simpler means. Starting from the same symbol-level segmentation from the previous methods, a VGG16 classification model is trained with the task of providing the label of the document a symbol comes from. Training is performed on a subset of the documents only. The task consists on classifying any new incoming document samples with labels from the original and count the times each document is classified as the second choice for the model. The assumption is that the second option can be interpreted as the easiest to confuse class with respect to the original document. Using this paradigm with the unseen classes, the results are shown in Table \ref{tab:martin} (after normalising by the number of samples of each input document).

\begin{table}[ht]
\caption{Similarity results using the classification method. Rows show the origin of the input documents and columns show what training document they are classified as. In bold, the preferred in-domain class for every input document.}
\label{tab:martin}
\centering
\resizebox{\columnwidth}{!}{%
\begin{tabular}{@{}lccccccc@{}}
\toprule
\diagbox[width=6em]{Test}{Train}& $\quad$ Borg $\quad$  & Copiale & Vatican 1 & Vatican 2 & Vatican 3 & Vatican 6 & Vatican 7 \\ \midrule
\multicolumn{1}{l}{BNF}           & 0.176 & 0.011   & 0         & \textbf{0.344}     & 0.327     & 0.141     & 0.001     \\
\multicolumn{1}{l}{Maximilian II $\quad$} & 0.122 & 0.085   & 0.026     & \textbf{0.349}     & 0.103     & 0.257     & 0.059     \\
\multicolumn{1}{l}{Ramanacoil}    & \textbf{0.335} & 0.034   & 0.002     & 0.243     & 0.244     & 0.137     & 0.005     \\
\multicolumn{1}{l}{Runicus}       & 0.002 & \textbf{0.744}   & 0.001     & 0.098     & 0.002     & 0.107     & 0.045     \\
\multicolumn{1}{l}{Zodiac}        & 0.02  & \textbf{0.747}   & 0         & 0.098     & 0.002     & 0.111     & 0.022     \\ \bottomrule
\end{tabular}}
\end{table}

Some aspects observed in the previous experiments such as the affinities of Zodiac with other ciphers are replicated here. Results for other ciphers are nevertheless not aligned with those from the proposed CSI metric or the clustering algorithm from the Baseline. Overall, this idea seems to have difficulties in capturing nuanced similarities between ciphers -- in most cases, most of the mass of samples falls on the same secondary class, from which a similarity metric cannot be extrapolated.


\section{Conclusions}
\label{sec:conc}


This work presents the CSI metric, a novel way of comparing pairs of ciphered documents in order to find whether they share alphabets with no prior knowledge about their contents. We have shown this method is capable of analysing ciphers containing both variations of the same alphabet as well as unique alphabets. We have also shown that this method improves upon previous exploratory works \cite{chen2021unsupervised} by giving results in a broader range that are more stable to the origin of the features and partially writing style. Finally, we have also used a simple secondary method to address the question of identifying the closest document in a known collection given a new unknown one.

There are some possibilities to improve on this work. Firstly, we plan to incorporate the expertise of a historical paleographer to better evaluate the strengths and weaknesses of the method and thus improve it. In particular, it would be interesting to compare how similar does an expert perceive two ciphers to be against what our method generates. Secondly, the fact that the symbol segmentation system is fully unsupervised makes it very flexible, but it also induces some assumptions that may not hold for all scripts (e.g. the degree of overlap between symbols is assumed to be low). Some outliers are fed into the model as a result. Thus, improving the segmentation algorithm is another potential road for improvement of the overall method.

\section*{Acknowledgments}

We thank Béata Megyesi, Crina Tudor, Michelle Waldispühl, Nils Kopal and George Lasry for providing the manuscript images. This work has been partially supported by the Swedish Research Council (grant 2018-06074, DECRYPT), the Spanish projects PID2021-126808OB-I00 (GRAIL) and CNS2022-135947 (DOLORES) from the Ministerio de Ciencia e Innovación, the Departament de Cultura of the Generalitat de Catalunya, and the CERCA Program / Generalitat de Catalunya. Pau Torras is funded by the Spanish FPU Grant FPU22/00207. Adrià Molina is funded with the PRE2022-101575 grant provided by MCIN / AEI / 10.13039 / 501100011033 and by the European Social Fund (FSE+).

\bibliographystyle{splncs04}
\bibliography{bibliography}

\begin{thebibliography}{10}
\providecommand{\url}[1]{\texttt{#1}}
\providecommand{\urlprefix}{URL }
\providecommand{\doi}[1]{https://doi.org/#1}

\bibitem{brito1997connectivity}
Brito, M.R., Ch{\'a}vez, E.L., Quiroz, A.J., Yukich, J.E.: Connectivity of the mutual k-nearest-neighbor graph in clustering and outlier detection. Statistics \& Probability Letters  \textbf{35}(1),  33--42 (1997)

\bibitem{chen2021unsupervised}
Chen, J., Souibgui, M.A., Forn{\'e}s, A., Megyesi, B.: Unsupervised alphabet matching in historical encrypted manuscript images. In: International Conference on Historical Cryptology. pp. 34--37 (2021)

\bibitem{christlein2017unsupervised}
Christlein, V., Gropp, M., Fiel, S., Maier, A.: Unsupervised feature learning for writer identification and writer retrieval. In: 2017 14th IAPR international conference on document analysis and recognition (ICDAR). vol.~1, pp. 991--997. IEEE (2017)

\bibitem{deng2009imagenet}
Deng, J., Dong, W., Socher, R., Li, L.J., Li, K., Fei-Fei, L.: Imagenet: A large-scale hierarchical image database. In: 2009 IEEE conference on computer vision and pattern recognition. pp. 248--255. Ieee (2009)

\bibitem{dinnissen2021island}
Dinnissen, J., Kopal, N.: Island ramanacoil a bridge too far. a dutch ciphertext from 1674. In: International Conference on Historical Cryptology. pp. 48--57 (2021)

\bibitem{dosovitskiy2021image}
Dosovitskiy, A., Beyer, L., Kolesnikov, A., Weissenborn, D., Zhai, X., Unterthiner, T., Dehghani, M., Minderer, M., Heigold, G., Gelly, S., Uszkoreit, J., Houlsby, N.: An image is worth 16x16 words: Transformers for image recognition at scale. arXiv preprint arXiv:2010.11929  (2021)

\bibitem{fischer2011transcription}
Fischer, A., Frinken, V., Forn{\'e}s, A., Bunke, H.: Transcription alignment of latin manuscripts using hidden markov models. In: Proceedings of the 2011 Workshop on Historical Document Imaging and Processing. pp. 29--36 (2011)

\bibitem{fischer2012lexicon}
Fischer, A., Keller, A., Frinken, V., Bunke, H.: Lexicon-free handwritten word spotting using character hmms. Pattern recognition letters  \textbf{33}(7),  934--942 (2012)

\bibitem{girvan2002community}
Girvan, M., Newman, M.E.: Community structure in social and biological networks. Proceedings of the national academy of sciences  \textbf{99}(12),  7821--7826 (2002)

\bibitem{graves2006connectionist}
Graves, A., Fern{\'a}ndez, S., Gomez, F., Schmidhuber, J.: Connectionist temporal classification: labelling unsegmented sequence data with recurrent neural networks. In: Proceedings of the 23rd international conference on Machine learning. pp. 369--376 (2006)

\bibitem{openclip_github}
Ilharco, G., Wortsman, M., Wightman, R., Gordon, C., Carlini, N., Taori, R., Dave, A., Shankar, V., Namkoong, H., Miller, J., Hajishirzi, H., Farhadi, A., Schmidt, L.: Openclip (Jul 2021). \doi{10.5281/zenodo.5143773}, \url{https://doi.org/10.5281/zenodo.5143773}

\bibitem{keserwani2019zero}
Keserwani, P., De, K., Roy, P.P., Pal, U.: Zero shot learning based script identification in the wild. In: 2019 international conference on document analysis and recognition (ICDAR). pp. 987--992. IEEE (2019)

\bibitem{nils2022deciphering}
Kopal, N., Waldispühl, M.: Deciphering three diplomatic letters sent by maximilian ii in 1575. Cryptologia  \textbf{46}(2),  103--127 (2022). \doi{10.1080/01611194.2020.1858370}, \url{https://doi.org/10.1080/01611194.2020.1858370}

\bibitem{leierzopf2021massive}
Leierzopf, E., Kopal, N., Esslinger, B., Lampesberger, H., Hermann, E.: A massive machine-learning approach for classical cipher type detection using feature engineering. In: International conference on historical cryptology. pp. 111--120 (2021)

\bibitem{leierzopf2021detection}
Leierzopf, E., Mikhalev, V., Kopal, N., Esslinger, B., Lampesberger, H., Hermann, E.: Detection of classical cipher types with feature-learning approaches. In: Data Mining: 19th Australasian Conference on Data Mining, AusDM 2021, Brisbane, QLD, Australia, December 14-15, 2021, Proceedings 19. pp. 152--164. Springer (2021)

\bibitem{liu2012noisy}
Liu, H., Zhang, S.: Noisy data elimination using mutual k-nearest neighbor for classification mining. Journal of Systems and Software  \textbf{85}(5),  1067--1074 (2012)

\bibitem{long2022towards}
Long, S., Qin, S., Panteleev, D., Bissacco, A., Fujii, Y., Raptis, M.: Towards end-to-end unified scene text detection and layout analysis. In: Proceedings of the IEEE/CVF Conference on Computer Vision and Pattern Recognition. pp. 1049--1059 (2022)

\bibitem{lowe2004distinctive}
Lowe, D.G.: Distinctive image features from scale-invariant keypoints. International journal of computer vision  \textbf{60},  91--110 (2004)

\bibitem{megyesi2020decryption}
Megyesi, B., Esslinger, B., Forn{\'e}s, A., Kopal, N., L{\'a}ng, B., Lasry, G., Leeuw, K.d., Pettersson, E., Wacker, A., Waldisp{\"u}hl, M.: Decryption of historical manuscripts: the decrypt project. Cryptologia  \textbf{44}(6),  545--559 (2020)

\bibitem{lucas_hermann_negri_2017_887917}
Negri, L.H., Vestri, C.: lucashn/peakutils: v1.1.0 (Sep 2017). \doi{10.5281/zenodo.887917}, \url{https://doi.org/10.5281/zenodo.887917}

\bibitem{nicolau2016visual}
Nicolaou, A., Bagdanov, A.D., Gomez{-}Bigorda, L., Karatzas, D.: Visual script and language identification. CoRR  \textbf{abs/1601.01885} (2016), \url{http://arxiv.org/abs/1601.01885}

\bibitem{opencv_thresholding_tutorial}
{OpenCV Team}: Image thresholding (06/05/2024), \url{https://docs.opencv.org/4.x/d7/d4d/tutorial_py_thresholding.html}

\bibitem{ozaki2011using}
Ozaki, K., Shimbo, M., Komachi, M., Matsumoto, Y.: Using the mutual k-nearest neighbor graphs for semi-supervised classification on natural language data. In: Proceedings of the fifteenth conference on computational natural language learning. pp. 154--162 (2011)

\bibitem{peratello2020codex}
Peratello, P., et~al.: Codex runicus (am 28 8vo): A pilot project for encoding a runic manuscript. Umanistica Digitale  \textbf{9},  155--169 (2020)

\bibitem{radford2021learning}
Radford, A., Kim, J.W., Hallacy, C., Ramesh, A., Goh, G., Agarwal, S., Sastry, G., Askell, A., Mishkin, P., Clark, J., Krueger, G., Sutskever, I.: Learning transferable visual models from natural language supervision. arXiv preprint:2103.00020  (2021)

\bibitem{romero2013esposalles}
Romero, V., Forn{\'e}s, A., Serrano, N., S{\'a}nchez, J.A., Toselli, A.H., Frinken, V., Vidal, E., Llad{\'o}s, J.: The esposalles database: An ancient marriage license corpus for off-line handwriting recognition. Pattern Recognition  \textbf{46}(6),  1658--1669 (2013)

\bibitem{schuhmann2022laion5b}
Schuhmann, C., Beaumont, R., Vencu, R., Gordon, C., Wightman, R., Cherti, M., Coombes, T., Katta, A., Mullis, C., Wortsman, M., Schramowski, P., Kundurthy, S., Crowson, K., Schmidt, L., Kaczmarczyk, R., Jitsev, J.: Laion-5b: An open large-scale dataset for training next generation image-text models. arXiv preprint:2210.08402  (2022)

\bibitem{shi2016script}
Shi, B., Bai, X., Yao, C.: Script identification in the wild via discriminative convolutional neural network. Pattern Recognition  \textbf{52},  448--458 (2016)

\bibitem{simonyan2015deep}
Simonyan, K., Zisserman, A.: Very deep convolutional networks for large-scale image recognition. arXiv preprint arXiv:1409.1556  (2015)

\bibitem{souibgui2022few}
Souibgui, M.A., Forn{\'e}s, A., Kessentini, Y., Megyesi, B.: Few shots are all you need: a progressive learning approach for low resource handwritten text recognition. Pattern Recognition Letters  \textbf{160},  43--49 (2022)

\bibitem{souibgui2023evaluation}
Souibgui, M.A., Torras, P., Chen, J., Forn{\'e}s, A.: An evaluation of handwritten text recognition methods for historical ciphered manuscripts. In: Proceedings of the 7th International Workshop on Historical Document Imaging and Processing. pp. 7--12 (2023)

\bibitem{terrades2022flexible}
Terrades, O.R., Berenguel, A., Gil, D.: A flexible outlier detector based on a topology given by graph communities. Big Data Research  \textbf{29},  100332 (2022)

\bibitem{vaswani2017attention}
Vaswani, A., Shazeer, N., Parmar, N., Uszkoreit, J., Jones, L., Gomez, A.N., Kaiser, {\L}., Polosukhin, I.: Attention is all you need. Advances in neural information processing systems  \textbf{30} (2017)

\bibitem{yin2019decipherment}
Yin, X., Aldarrab, N., Megyesi, B., Knight, K.: Decipherment of historical manuscript images. In: 2019 International Conference on Document Analysis and Recognition (ICDAR). pp. 78--85. IEEE (2019)

\bibitem{zheng2017sift}
Zheng, L., Yang, Y., Tian, Q.: Sift meets cnn: A decade survey of instance retrieval. IEEE transactions on pattern analysis and machine intelligence  \textbf{40}(5),  1224--1244 (2017)

\end{thebibliography}

\end{document}